# Characterising Research Areas in the field of AI
## *Temi di ricerca caratterizzanti nel campo dell'IA*

Alessandra Belfiore[1], Angelo Salatino[2], Francesco Osborne[3]

**Abstract** Interest in Artificial Intelligence (AI) continues to grow rapidly, hence it is crucial to support researchers and organisations in understanding where AI research is heading. In this study, we conducted a bibliometric analysis on 257K articles in AI, retrieved from OpenAlex. We identified the main conceptual themes by performing clustering analysis on the co-occurrence network of topics. Finally, we observed how such themes evolved over time. The results highlight the growing academic interest in research themes like deep learning, machine learning, and internet of things.

**Abstract** *L'interesse nell'intelligenza artificiale (AI) continua a crescere rapidamente, per questo è importante aiutare ricercatori e organizzazioni nel comprendere dove si sta dirigendo la ricerca in AI. In questo studio, abbiamo eseguito un'analisi bibliometrica su 275 mila articoli di ricerca in AI, scaricati da OpenAlex. Abbiamo identificato i principali temi concettuali eseguendo un'analisi dei gruppi sulla rete delle co-occorrenze dei topic. Infine, abbiamo osservato come questi temi si sviluppano nel tempo. I risultati mostrano un crescente interesse accademico nei temi di ricerca come deep learning, machine learning, e internet of things.*

## 1 Introduction

Interest in Artificial Intelligence (AI) continues to grow rapidly, hence it is crucial to support researchers and organisations with novel ways of exploring the scientific landscape as they can take informed decisions.

In this paper, we present a bibliometric analysis on the recent trends in AI. In particular, we initially downloaded 257K papers in the field of AI from OpenAlex, from the 1990 to February 2022, and we associated them with research topics in the Computer Science Ontology (CSO), the largest ontology of research topics in the field of Computer Science. Then, we organised all the documents in 7 periods based on the publishing year. In each time period, we first identified conceptual themes (i.e., clusters of topics) representing research areas and then we computed the Callon's indices of density and centrality. These indices allowed us to determine whether the themes are motor, niche, basic, and emerging (or declining). Finally, we mapped the similar themes across the different timeframes and analysed how they developed over time: e.g., before they started being niche and after became motor.

---

[1] Alessandra Belfiore, Università della Campania Luigi Vanvitelli; email: alessandra.belfiore@unicampania.it

[2] Angelo Salatino, The Open University; email: angelo.salatino@open.ac.uk

[3] Francesco Osborne, The Open University; email: francesco.osborne@open.ac.uk

In this analysis, we identified eight themes experiencing a significant shift, which we explained with actual events happened in the field of Artificial Intelligence, such as the Deep Learning revolution and the emergence of IoT.

The remainder of the paper is organised as follows. In Section 2, we present our dataset and methodology. In Section 3, we present our results. Finally, Section 4 concludes the paper, outlining future directions.

## 2  Material and Methods

To perform this analysis, we first downloaded the research papers in the field of AI, from OpenAlex[1], a recently launched scholarly dataset. Then, we run the CSO Classifier on the papers metadata (title and abstracts) to extract their relevant research topics, and finally we run the thematic analysis to assess how the various themes evolved over time.

We used openalexR (Aria, (2022)) to retrieve all papers having "artificial intelligence", "machine learning", "deep learning" and "data science" either in titles or abstracts, published during the period 1990 to February 2022 inclusive, resulting in 257K research papers.

We extracted the relevant topics from all the research documents with the CSO Classifier[2] (Salatino et al., (2019)), a tool that takes in input the text of a research paper (title, abstract and keyword) and returns a selection of research topics drawn from the Computer Science Ontology (Salatino et al., (2018)).

After associating each document with its relevant research topics, we split the corpus in 7 timeframes. The first six timeframes are of 5 years each (1990-94, 1995-99, up to 2015-19), the last timeframe goes from 2020 to 2022. In each timeframe, we identified and characterised the different conceptual themes and then we observed how they evolved over time. For instance, we may want to detect when and whether a theme became highly relevant and well developed.

As a first step, in each timeframe, we created the topic co-occurrence network using the topics returned by the CSO Classifier. The topic co-occurrence network is a fully weighted graph describing the interaction between topics. In this graph, nodes (i.e., topics) are linked together by undirected arcs to describe the extent of their co-occurrence. The node weight represents the number of publications that a topic has received in such timeframe, and the link weight is equal to the number of papers the two topics appeared together in the same period.

---

[1] OpenAlex - https://openalex.org
[2] The CSO Classifier - https://github.com/angelosalatino/cso-classifier

We applied the Louvain community detection algorithm (Blondel et al., (2008)) on these networks to extract clusters of topics (i.e., conceptual themes). For this, we leveraged Bibliometrix (Aria and Cuccurullo, (2017)), which is an open-source tool developed in R for quantitative research in scientometrics and bibliometrics. Specifically, as parameters we set 1000 topics, with the minimum cluster frequency set to 5.

We computed the Callon's centrality and density indices on the resulting clusters to respectively measure the relevance and the degree of development of the theme (Callon et al., (1991); Aria et al., (2020)).

Based on the values of both centrality and density, each cluster has been classified according to four themes: i) motor, ii) basic, iii) emerging or declining, and iv) niche (He, (1999); Cahlik, (2000)). The *motor themes* are highly relevant and well developed in research, as they have levels of centrality and density above average. The *basic themes* are low developed in research but relevant themes, displaying low levels of density and high levels of centrality. The *emerging or declining themes* have the lowest levels of density and centrality. This occurs in two moments of their life: when they either emerge or decline. The distinction between emerging or declining themes can be understood only by comparing their evolution over time. Finally, the *niche themes* are highly developed, but they are developed by a small niche of researchers, displaying density above average and centrality below average.

After determining the class of all the extracted conceptual themes in the 7 timeframes, we mapped the similar ones appearing in multiple timeframes. Two themes in consecutive time periods have been mapped if they had the same top-3 topics. We also mapped together two themes that differed of just one topic, but at the condition that the unmatched topic was among the top-5 topics of the other theme.

We used this mapping to analyse the most significant shifts in this space.

## 3   Results

This section presents the main results of our analysis on the evolution of conceptual themes in the considered seven periods.

The Bibliometrix tool extracted from the topic co-occurrence networks 6-9 conceptual themes ($7.42 \pm 0.97$) in each period. We observed that some themes appeared just once or in two consecutive timeframes, whereas some others appeared in multiple time periods. To this end, we only mapped the clusters that recurred in three or more consecutive timeframes so to attain a better understanding of their life trajectory. Altogether, we identified eight recurring clusters portraying interesting dynamics. In Table 2, we report the eight themes, identified by their highly representative topic, and their classification over the seven time periods based on Callon's centrality and density.

It is worth pointing out that the four-themed classification and the life trajectory is contextualised to the whole cluster and not just its highly representative topic. The context surrounding the first conceptual theme is about expert systems, intelligent systems, and more in general symbolic AI which has experienced a steady decline in the past decades as an increasing number of researchers have shifted their focus to probabilistic AI. The second conceptual theme is related to machine learning and includes relevant research areas such as supervised machine learning and neural networks. According to the data, initially it was highly relevant for the community (motor), but in the following it lost some momentum, until its resurgence in recent years, also thanks to the availability of more powerful machines that can handle large machine learning models.

**Table 1.** Recurring conceptual themes identified in the seven temporal times. In the theme column, we report only the most representative topic.

| Theme | 1990-94 | 1995-99 | 2000-04 | 2005-09 | 2010-14 | 2015-19 | 2020-22 |
|---|---|---|---|---|---|---|---|
| **expert systems** | decline | decline | decline | decline | decline | decline | decline |
| **machine learning** | motor | basic | basic | decline | basic | basic | motor |
| **reasoning** | basic | motor | motor | - | - | - | - |
| **data mining** | - | - | niche | motor | motor | - | - |
| **genetic algorithms** | - | motor | motor | motor | decline | - | - |
| **sensors** | - | - | - | emerging | motor | motor | motor |
| **robots** | niche | basic | decline | motor | - | - | - |
| **deep learning** | - | - | - | - | niche | motor | motor |

The third theme includes reasoning, multiagent systems, semantics, logic programming, and intelligent agents, outlining the multi-agent era. The theme went off the radar from the 2005 onward. This finding is confirmed by previous bibliometric analysis (Osborne, et al., (2014)) and discussed by some articles at that time. For instance, a 2007 editorial titled "Where are all the Intelligent Agents?" (Hendler, (2007)) suggested that the role of agent research in Semantic Web community was not as strong as envisaged in the original 2001 vision.

The fourth theme, mostly associated with data mining, shows the emergence of the big data era and the applications of knowledge discovery. We do not have data in the last two time periods due to the sensitivity of the algorithm in returning the relevant clusters. In the future, we plan to run a deeper analysis investigating a high number of clusters per period.

The fifth theme includes topics like genetic algorithms, adaptive algorithms, optimization, and optimization problems, which had their culmination in the decade 2000-10.

The sixth theme identified by sensors shows the life trajectory of the internet of things, emerged in 2005-09 and currently highly relevant and well developed.

The seventh theme shows the application of robots in control systems, including sensors, switching control, and process control, which starts being niche and goes through a decline up to 2004. From 2005 onwards there is a paradigm shift in which the theme of robots includes neural network architectures, making it highly relevant.

Finally, the eighth theme is centred around deep learning including pattern recognition, neural networks, convolutional neural networks, and deep belief network, which was niche in 2010, but from 2015 onwards it started being highly relevant and well developed.

## 4 Conclusions and Future Work

In this paper, we performed a bibliometric analysis in the domain of Artificial Intelligence, and we observed how conceptual themes evolved over time.

We identified eight conceptual themes experiencing significant shift, signalling how the AI field is highly dynamic, and we are confident this will continue in the future as the attention to AI is growing. Specifically, we observed the development over time of specific themes like Deep Learning, IoT, Robotics, Machine Learning, Genetic Algorithms and then we provided an explanation for such shift based on actual events happened in the field of AI.

The findings of this study have to be seen in light of some limitations. Both the thematic evolution and the conceptual themes can be affected by the selection of papers we retrieved from OpenAlex.

For the future, we plan to work on multiple fronts. First, we would like to increase the sensitivity of the Bibliometrix tool to return more than 9 clusters of topics in each timeframe. This would enable us to perform a more comprehensive analysis. Second, we plan to analyse the whole Computer Science, so to be able to characterise a larger pool of conceptual themes. Third, we plan to perform a qualitative analysis to understanding how the four-themed classification relates to the Khun's phases of scientific revolution (Kuhn, (1970)). Finally, we plan to analyse the patterns of Callon's centrality and density to understand whether they have predictive power to forecast how novel topics will develop in the upcoming years.